\documentclass[11pt,a4paper]{article}
\usepackage[hyperref]{acl2017}
\usepackage{times}
\usepackage{latexsym}

\usepackage{url}

\usepackage{algorithm} %format of the algorithm
\usepackage{algorithmic} %format of the algorithm
\usepackage{multirow} %multirow for format of table
\usepackage{amsmath}
\usepackage{xcolor}
\usepackage{epsfig}
\usepackage{array}
\usepackage{bbding}
\usepackage{threeparttable}
\usepackage{CJK}
\aclfinalcopy % Uncomment this line for the final submission
\def\aclpaperid{3224} %  Enter the acl Paper ID here

\newcommand*{\affaddr}[1]{#1} % No op here. Customize it for different styles.
\newcommand*{\affmark}[1][*]{\textsuperscript{#1}}
\newcommand*{\email}[1]{\texttt{#1}}

\title{Improving Semantic Relevance for Sequence-to-Sequence Learning of Chinese Social Media Text Summarization}

\author{%
Shuming Ma\affmark[1,2], Xu Sun\affmark[1,2], Jingjing Xu\affmark[1,2], Houfeng Wang\affmark[1,2], Wenjie Li\affmark[3], Qi Su\affmark[4]\\
\affaddr{\affmark[1]MOE Key Laboratory of Computational Linguistics, Peking University}\\
\affaddr{\affmark[2]School of Electronics Engineering and Computer Science, Peking University}\\
\affaddr{\affmark[3]Department of Computing, The Hong Kong Polytechnic University}\\
\affaddr{\affmark[4]School of Foreign Languages, Peking University}\\
\email{\{shumingma, xusun, jingjingxu, wanghf, sukia\}@pku.edu.cn}\\
\email{cswjli@comp.polyu.edu.hk}
}

\date{}

\begin{document}
\begin{CJK}{UTF8}{gbsn}
\maketitle
\begin{abstract}
Current Chinese social media text summarization models are based on an encoder-decoder framework. Although its generated summaries are similar to source texts literally, they have low semantic relevance. In this work, our goal is to improve semantic relevance between source texts and summaries for Chinese social media summarization. We
introduce a Semantic Relevance Based neural model to encourage high semantic similarity between texts and summaries. In our model, the source text is represented by a gated attention encoder, while the summary representation is produced by a decoder. Besides, the similarity score between the representations is maximized during training. Our experiments show that the proposed model outperforms baseline systems on a social media corpus.
\end{abstract}

\section{Introduction}

Text summarization is to produce a brief summary of the main ideas of the text. For long and normal documents, extractive summarization achieves satisfying performance by selecting a few sentences from source texts~\cite{extra04,extra10,discourse}. However, it does not apply to Chinese social media text summarization, where texts are comparatively short and often full of noise. Therefore, abstractive text summarization, which is based on encoder-decoder framework, is a better choice~\cite{abs,lcsts}.

For extractive summarization, the selected sentences often have high semantic relevance to the text. However, for abstractive text summarization, current models tend to produce grammatical and coherent summaries regardless of its semantic relevance with source texts. Figure~\ref{fig1} shows that the summary generated by a current model (RNN encoder-decoder) is similar to the source text literally, but it has low semantic relevance.

%%%%%%%%%%%%%%%%%%%%%%%%%%%%%%%%%%%%%%%%%%%%%%%%%%
\begin{figure}[tb]
\centering
\begin{tabular}{|l|}
\hline
Text: 昨晚，\underline{中联航空}成都飞北京一架航班\\
被发现有\underline{多人}吸烟。后因天气原因，飞机\\
备降太原\underline{机场}。有乘客要求重新安检，机\\
长决定继续飞行，引起机组人员与未吸烟\\
乘客冲突。\\
Last night, \underline{several people} were caught to smo-\\
ke on a flight of \underline{China United Airlines} from \\
Chendu to Beijing. Later the flight temporari-\\
ly landed on Taiyuan \underline{Airport}. Some passeng-\\
ers asked for a security check but were denied \\
by the captain, which led to a collision betwe-\\
en crew and passengers. \\
\\
RNN: \underline{中联航空机场}发生爆炸致\underline{多人}死亡。\\
\underline{China United Airlines} exploded in the \underline{airport}, \\
leaving \underline{several people} dead.  \\
\\
Gold: 航班多人吸烟机组人员与乘客冲突。  \\
Several people smoked on a flight which led\\
to a collision between crew and passengers. \\
\hline
\end{tabular}
\caption{An example of RNN generated summary. It has high similarity to the text literally, but low semantic relevance.
}\label{fig1}
\vspace{-0.1in}
\end{figure}
%%%%%%%%%%%%%%%%%%%%%%%%%%%%%%%%%%%%%%%%%%%%%%%%%%

In this work, our goal is to improve the semantic relevance between source texts and generated summaries for Chinese social media text summarization. To achieve this goal, we propose a Semantic Relevance Based neural model. In our model, a similarity evaluation component is introduced to measure the relevance of source texts and generated summaries. During training, it maximizes the similarity score to encourage high semantic relevance between source texts and summaries. The representation of source texts is produced by an encoder, while that of summaries is computed by a decoder. We introduce a gated attention encoder to better represent the source text. Besides, our decoder generates summaries and provide the summary representation. Experiments show that our proposed model has better performance than baseline systems on the social media corpus.

\section{Background: Chinese Abstractive Text Summarization}
\label{background}

Current Chinese social media text summarization model is based on encoder-decoder framework. Encoder-decoder model is able to compress source texts $x=\{x_1,x_2,...,x_N\}$ into continuous vector representation with an encoder, and then generate the summary $y=\{y_1,y_2,...,y_M\}$ with a decoder. In the previous work~\cite{lcsts}, the encoder is a bi-directional gated recurrent neural network, which maps source texts into sentence vector $\{h_1,h_2,...,h_N\}$. The decoder is a uni-directional recurrent neural network, which produces the distribution of output words $y_t$ with previous hidden state $s_{t-1}$ and word $y_{t-1}$:
\begin{equation}\label{decoder}
  p(y_t|x)=softmax{f(s_{t-1},y_{t-1})}
\end{equation}
where $f$ is recurrent neural network output function, and $s_0$ is the last hidden state of encoder $h_N$.

Attention mechanism is introduced to better capture context information of source texts~\cite{attention}. Attention vector $c_t$ is represented by the weighted sum of encoder hidden states:
\begin{equation}\label{attention1}
  c_t=\sum_{i=1}^{N}{\alpha_{ti}h_{i}}
\end{equation}
\begin{equation}\label{attention2}
  \alpha_{ti}=\frac{e^{g(s_{t},h_{i})}}{\sum_{j=1}^{N}{e^{g(s_{t},h_{j})}}}
\end{equation}
where $g(s_{t},h_{i})$ is a relevant score between decoder hidden state $s_t$ and encoder hidden state $h_i$. When predicting an output word, the decoder takes account of attention vector, which contains the alignment information between source texts and summaries.

%%%%%%%%%%%%%%%%%%%%%%%%%%%%%%%%%%%%%%%%%%%%%%%%%%
\begin{figure}[tb]
\centering
\begin{tabular}{@{}c@{}@{}c@{}@{}c@{}@{}c@{}}

\epsfig{file=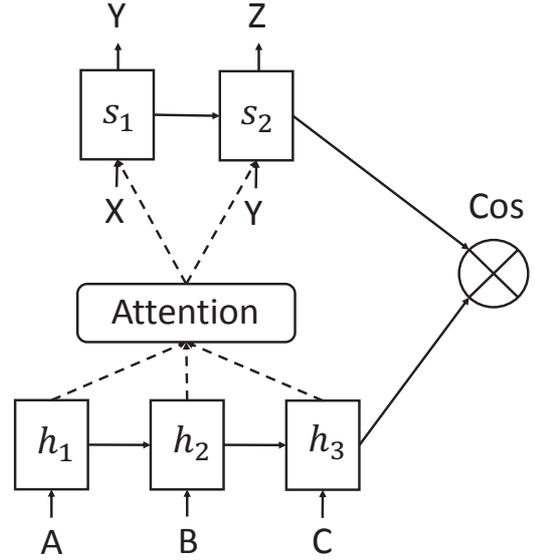,width=1.0\linewidth,clip=}

\end{tabular}
\caption{Our Semantic Relevance Based neural model. It consists of decoder (above), encoder (below) and cosine similarity function.
}\label{fig2}
\vspace{-0.1in}
\end{figure}
%%%%%%%%%%%%%%%%%%%%%%%%%%%%%%%%%%%%%%%%%%%%%%%%%%

\section{Proposed Model}

Our assumption is that source texts and summaries have high semantic relevance, so our proposed model encourages high similarity between their representations. Figure~\ref{fig2} shows our proposed model. The model consists of three components: encoder, decoder and a similarity function. The encoder compresses source texts into semantic vectors, and the decoder generates summaries and produces semantic vectors of the generated summaries. Finally, the similarity function evaluates the relevance between the sematic vectors of source texts and generated summaries. Our training objective is to maximize the similarity score so that the generated summaries have high semantic relevance with source texts.

\subsection{Text Representation}

There are several methods to represent a text or a sentence, such as mean pooling of RNN output or reserving the last state of RNN. In our model, source text is represented by a gated attention encoder~\cite{humanreading}. Every upcoming word is fed into a gated attention network, which measures its importance. The gated attention network outputs the important score with a feedforward network. At each time step, it inputs a word vector $e_t$ and its previous context vector $h_t$, then outputs the score $\beta_t$. Then the word vector $e_t$ is multiplied by the score $\beta_t$, and fed into RNN encoder. We select the last output $h_N$ of RNN encoder as the semantic vector of the source text $V_{t}$.

A natural idea to get the semantic vector of a summary is to feed it into the encoder as well. However, this method wastes much time because we encode the same sentence twice. Actually, the last output $s_M$ contains information of both source text and generated summaries. We simply compute the semantic vector of the summary by subtracting $h_N$ from $s_M$:
\begin{equation}
  V_{s}=s_M-h_N
\end{equation}
Previous work has proved that it is effective to represent a span of words without encoding them once more~\cite{wang16}.

\subsection{Semantic Relevance}

Our goal is to compute the semantic relevance of source text and generated summary given semantic vector $V_t$ and $V_s$. Here, we use cosine similarity to measure the semantic relevance, which is represented with a dot product and magnitude:
\begin{equation}
  cos(V_s,V_t)=\frac{V_{s} \cdot V_{t}}{\Vert V_s \Vert \Vert V_t \Vert}
\end{equation}
Source text and summary share the same language, so it is reasonable to assume that their semantic vectors are distributed in the same space. Cosine similarity is a good way to measure the distance between two vectors in the same space.

\subsection{Training}

Given the model parameter $\theta$ and input text $x$, the model produces corresponding summary $y$ and semantic vector $V_s$ and $V_t$. The objective is to minimize the loss function:
\begin{equation}
  L=-p(y|x;\theta)-\lambda cos(V_s,V_t)
\end{equation}
where $p(y|x;\theta)$ is the conditional probability of summaries given source texts, and is computed by the encoder-decoder model. $cos(V_s,V_t)$ is cosine similarity of semantic vectors $V_s$ and $V_t$. This term tries to maximize the semantic relevance between source input and target output.

%%%%%%%%%%%%%%%%%%%%%%%%%%%%%%%%%%%%%%%%%%%%%%%%%%
%\begin{figure}[tb]
%\centering
%\begin{tabular}{@{}c@{}@{}c@{}@{}c@{}@{}c@{}}

%\epsfig{file=pic_2.eps,width=0.8\linewidth,clip=}

%\end{tabular}
%\caption{A gated attention encoder.
%}\label{fig3}
%\vspace{-0.1in}
%\end{figure}
%%%%%%%%%%%%%%%%%%%%%%%%%%%%%%%%%%%%%%%%%%%%%%%%%%

\section{Experiments}

In this section, we present the evaluation of our model and show its performance on a popular social media corpus. Besides, we use a case to explain the semantic relevance between generated summary and source text.

\subsection{Dataset}

Our dataset is Large Scale Chinese Short Text Summarization Dataset (LCSTS), which is constructed by Hu et al.~\shortcite{lcsts}. The dataset consists of more than 2.4 million text-summary pairs, constructed from a famous Chinese social media website called Sina Weibo\footnote{weibo.sina.com}. It is split into three parts, with 2,400,591 pairs in PART I, 10,666 pairs in PART II and 1,106 pairs in PART III. All the text-summary pairs in PART II and PART III are manually annotated with relevant scores ranged from 1 to 5, and we only reserve pairs with scores no less than 3. Following the previous work, we use PART I as training set, PART II as development set, and PART III as test set.

\subsection{Experiment Setting}

To alleviate the risk of word segmentation mistakes~\cite{Xu2016Dependency}, we use Chinese character sequences as both source inputs and target outputs. We limit the model vocabulary size to 4000, which covers most of the common characters. Each character is represented by a random initialized word embedding. We tune our parameter on the development set. In our model, the embedding size is 400, the hidden state size of encoder-decoder is 500, and the size of gated attention network is 1000. We use Adam optimizer to learn the model parameters, and the batch size is set as 32. The parameter $\lambda$ is 0.0001. Both the encoder and decoder are based on LSTM unit. Following the previous work~\cite{lcsts}, our evaluation metric is F-score of ROUGE: ROUGE-1, ROUGE-2 and ROUGE-L~\cite{rough}.

\subsection{Baseline Systems}

\noindent\textbf{RNN}. We denote RNN as the basic sequence-to-sequence model with bi-directional GRU encoder and uni-directional GRU decoder. It is a widely used language generated framework, so it is an important baseline.

\noindent\textbf{RNN context}. RNN context is a sequence-to-sequence framework with neural attention. Attention mechanism helps capture the context information of source texts. This model is a stronger baseline system.

\subsection{Results and Discussions}

%%%%%%%%%%%%%%%%%%%%%%%%%%%%%%%%%%%%%%%%%%%%%%%%%%%%%%%%%%%%%%
%\begin{table*}[tb]
%\centering
%\newcommand{\tabincell}[2]{\begin{tabular}{@{}#1@{}}#2\end{tabular}}
%\begin{tabular}{|l|c|c|c|c|c|c|c|c|c|}
%\hline
% \multirow{2}{*}{Model}  & \multicolumn{3}{c|}{ROUGE-1} & \multicolumn{3}{c|}{ROUGE-2} & \multicolumn{3}{c|}{ROUGE-L} \\
% \cline{2-10}
% & F & P & R & F & P & R & F & P & R \\
%\hline
%RNN (W)~\cite{lcsts}  &     17.7 & 13.8 & 27.0 & 8.5 & 6.6 & 13.4 &  15.8 & 12.3 & 24.3\\
%RNN (C)~\cite{lcsts}  &       21.5 & 17.6 & 31.1 & 8.9 & 7.2 & 13.4 &  18.6 & 15.2 & 27.1 \\
%RNN context (W)~\cite{lcsts} & 26.8 & 20.9 & \textbf{40.6} & 16.1 & 12.5 & \textbf{24.8} &  24.1 & 18.9 & \textbf{36.6} \\
%RNN context (C)~\cite{lcsts} & 29.9 & 27.9 & 35.4 & 17.4 & 16.1 & 21.0 &  27.2 & 25.4 & 32.3 \\
%\hline
%Our RNN-C & +C & 30.1 & 17.9 & 27.2 \\
%RNN context + SRB (C) & 32.1 & 30.7 & 35.4 & 18.9 & 17.6 & 22.1 & 29.2 & 28.0 & 32.3 \\
%+Attention (C) & \textbf{33.3} & \textbf{31.4} & 37.4  & \textbf{20.0} & \textbf{18.8} & 22.9 & \textbf{30.1} & \textbf{28.8} & 33.1 \\
%\hline
%\end{tabular}
%\caption{Results of our model and baseline systems. (W: Word level; C: Character level; F: F-score; P: Precision; R: Recall)} \label{table1}
%\end{table*}
%%%%%%%%%%%%%%%%%%%%%%%%%%%%%%%%%%%%%%%%%%%%%%%%%%%%%%%%%%%%%%

%%%%%%%%%%%%%%%%%%%%%%%%%%%%%%%%%%%%%%%%%%%%%%%%%%%%%%%%%%%%%%
\begin{table*}[tb]
\centering
\newcommand{\tabincell}[2]{\begin{tabular}{@{}#1@{}}#2\end{tabular}}
\begin{tabular}{|c|c|c|c|}
\hline
 Model  & ROUGE-1 & ROUGE-2 & ROUGE-L \\
\hline
RNN (W)~\cite{lcsts}  &     17.7  & 8.5  &  15.8 \\
RNN (C)~\cite{lcsts}  &       21.5  & 8.9  &  18.6  \\
RNN context (W)~\cite{lcsts} & 26.8  & 16.1  &  24.1  \\
RNN context (C)~\cite{lcsts} & 29.9 & 17.4  &  27.2  \\
\hline
%Our RNN-C & +C & 30.1 & 17.9 & 27.2 \\
RNN context + SRB (C) & 32.1  & 18.9  & 29.2  \\
+Attention (C) & \textbf{33.3}   & \textbf{20.0}  & \textbf{30.1} \\
\hline
\end{tabular}
\caption{Results of our model and baseline systems. Our models achieve substantial improvement of all ROUGE scores over baseline systems. (W: Word level; C: Character level).} \label{table1}
\end{table*}
%%%%%%%%%%%%%%%%%%%%%%%%%%%%%%%%%%%%%%%%%%%%%%%%%%%%%%%%%%%%%%

%%%%%%%%%%%%%%%%%%%%%%%%%%%%%%%%%%%%%%%%%%%%%%%%%%
\begin{figure}[tb]
\centering
\begin{tabular}{|l|}
\hline
Text:仔细一算，上海的互联网公司不乏成功\\
案例，但最终成为BAT一类巨头的几乎没有,\\
这也能解释为何纳税百强的榜单中鲜少互联\\
网公司的身影。有一类是被并购，比如：易\\
趣、土豆网、PPS、PPTV、一号店等；有一\\
类是数年偏安于细分市场。\\
With careful calculation, there are many succe-\\
ssful Internet companies in Shanghai, but few\\
of them becomes giant company like BAT. Th- \\
is is also the reason why few Internet compan-\\
ies are listed in top hundred companies of pay-\\
ing tax. Some of them are merged, such as Eb-\\
ay, Tudou, PPS, PPTV, Yihaodian and so on.\\
Others are satisfied with segment market for\\
years.\\
\\
Gold:为什么上海出不了互联网巨头？\\
Why Shanghai comes out no giant company?\\
\\
RNN context:上海的互联网巨头。\\
Shanghai's giant company.\\
\\
SRB:上海鲜少互联网巨头的身影。\\
Shanghai has few giant companies.\\
\hline
\end{tabular}
\caption{An Example of RNN generated summary on LCSTS corpus.
}\label{fig4}
\vspace{-0.01in}
\end{figure}
%%%%%%%%%%%%%%%%%%%%%%%%%%%%%%%%%%%%%%%%%%%%%%%%%%

\begin{table}[tb]
\centering
\newcommand{\tabincell}[2]{\begin{tabular}{@{}#1@{}}#2\end{tabular}}
\begin{tabular}{|c|c|c|c|c|}
\hline
 Model & level & R-1 & R-2 & R-L \\
\hline
RNN context& Word & 26.8 & 16.1 &  24.1 \\
\cline{2-5}
\cite{lcsts}& Char & 29.9 & 17.4  &  27.2 \\
\hline
COPYNET& Word & 35.0 & 22.3 & 32.0 \\
\cline{2-5}
\cite{copynet}& Char & 34.4 & 21.6  &  31.3 \\
\hline
this work & Char & 33.3 & 20.0 & 30.1 \\
\hline
\end{tabular}
\caption{Results of our model and state-of-the-art systems. COPYNET incorporates copying mechanism to solve out-of-vocabulary problem, so its has higher ROUGE scores. Our model does not incorporate this mechanism
currently. In the future work, we will implement this technic to further improve the performance. (Word: Word level; Char: Character level; R-1: F-score of ROUGE-1; R-2: F-score of ROUGE-2; R-L: F-score of ROUGE-L)} \label{table2}
\end{table}

We compare our model with above baseline systems, including RNN and RNN context. We refer to our proposed Semantic Relevance Based neural model as \textbf{SRB}.
Besides, SRB with a gated attention encoder is denoted as \textbf{+Attention}. Table~\ref{table1} shows the results of our models and baseline systems.
We can see SRB outperforms both RNN and RNN context in the F-score of ROUGE-1, ROUGE-2 and ROUGE-L. It concludes that SRB generates more key words and phrases.
With a gated attention encoder, SRB achieves a better performance with 33.3 F-score of ROUGE-1, 20.0 ROUGE-2 and 30.1 ROUGE-L. It shows that the gated attention reduces
noisy and unimportant information, so that the remaining information represents a clear idea of source text. The better representation of encoder leads to a
better semantic relevance evaluation by the similarity function. Therefore, SRB with gated attention encoder is able to generate summaries with high semantic
relevance to source text.

Figure~\ref{fig4} is an example to show the semantic relevance between the source text and the summary. It shows that the main idea of the source text is
about the reason why Shanghai has few giant company. RNN context produces ``Shanghai's giant companies'' which is literally similar to the source text, while
SRB generates ``Shanghai has few giant companies'', which is closer to the main idea in semantics. It concludes that SRB produces summaries with higher
semantic similarity to texts.

Table~\ref{table2} summarizes the results of our model and state-of-the-art systems. COPYNET has the highest socres, because it incorporates copying mechanism to deals with out-of-vocabulary word problem.
%In this paper, we focus more on the semantic relevance between the source texts and the summaries, and we have no processing with out-of-vocabulary words in our model.
%However, our model applies to most of sequence-to-sequence learning models, including COPYNET.
In this paper, we do not implement this mechanism in our model.
In the future work, we will try to incorporates copying mechanism to our model to solve the out-of-vocabulary problem.

\section{Related Work}

Abstractive text summarization has achieved successful performance thanks to the sequence-to-sequence model~\cite{seq2seq} and attention mechanism~\cite{attention}.
Rush et al.~\shortcite{abs} first used an attention-based encoder to compress texts and a neural network language decoder to generate summaries.
Following this work, recurrent encoder was introduced to text summarization, and gained better performance~\cite{rnnheadline,ras}. Towards Chinese texts, Hu et al.~\shortcite{lcsts}
built a large corpus of Chinese short text summarization. To deal with unknown word problem, Nallapati et al.~\shortcite{ibmsummarization} proposed a generator-pointer model
so that the decoder is able to generate words in source texts. Gu et al.~\shortcite{copynet} also solved this issue by incorporating copying mechanism.
Besides, Ayana et al.~\shortcite{minimum} proposes a minimum risk training method which optimizes the parameters with the target of rouge scores.

Our work is also related to neural attention model. Neural attention model is first proposed by Bahdanau et al.~\shortcite{attention}.
There are many other methods to improve neural attention model~\cite{mapattention,stanfordattention} and accelerate the training process~\cite{Sun2016Asynchronous}.

\section{Conclusion}

Our work aims at improving semantic relevance of generated summaries and source texts for Chinese social media text summarization.
Our model is able to transform the text and the summary into a dense vector, and encourage high similarity of their representation.
Experiments show that our model outperforms baseline systems, and the generated summary has higher semantic relevance.

\section*{Acknowledgements}

This work was supported in part by National High Technology Research and Development Program of China (863 Program, No. 2015AA015404),
and National Natural Science Foundation of China (No. 61673028). Xu Sun is the corresponding author of this paper.

%\bibliography{acl2017}
\bibliography{acl2017}
\bibliographystyle{acl_natbib}

\end{CJK}
\end{document}